# An Entropy Equation for Energy

Kieran Greer, Distributed Computing Systems, Belfast, UK.
http://distributedcomputingsystems.co.uk
Version 1.1

***Abstract -*** This paper describes an entropy equation, but one that should be used for measuring energy and not information. In relation to the human brain therefore, both of these quantities can be used to represent the stored information. The human brain makes use of energy efficiency to form its structures, which is likely to be linked to the neuron wiring. This energy efficiency can also be used as the basis for a clustering algorithm, which is described in a different paper. This paper is more of a discussion about global properties, where the rules used for the clustering algorithm can also create the entropy equation E = (mean * variance). This states that work is done through the energy released by the 'change' in entropy. The equation is so simplistic and generic that it can offer arguments for completely different domains, where the journey ends with a discussion about global energy properties in physics and beyond. A comparison with Einstein's relativity equation is made and also the audacious suggestion that a black hole has zero-energy inside.

## 1   Introduction

This paper explores the possibility that the human brain really does obey the laws of the Universe. It starts by introducing a self-organising system, based on brain economic principles. Implementation details of this are described in a different paper [7] that also describes a new agglomerative clustering procedure in detail. These rules were in fact, also used to produce a measure of cohesion [8], where this paper uses the same equation for a measure of entropy [4]. The clustering algorithm does not involve entropy but requires a much more detailed algorithmic procedure. In fact, the cohesion equation may not be very good at information clustering and the rules derived from the cognitive modelling research suggest that it is more akin to an energy comparison than an information one.  The form of the equation also bears a strong resemblance to Einstein's famous relativity equation [3], and so as part of this discussion, some comments about relativity, electromagnetism and black holes are made.

The rest of this paper is organised as follows: section 2 introduces the concepts used in this paper through a related work section. Section 3 describes the origin of the equation and how it relates to the human brain. Section 4 then extends the discussion to an equation about entropy and then to one about energy. Finally, section 5 ends with some conclusions on the work.

## 2    Related Work

The related work section is necessarily filled with references to Physics, as that is where the discussion in this paper ends. The topics include entropy [5], gravity [4], black holes [10] and even the general theory of relativity by Albert Einstein [3]. Entropic Gravity [12] might be particularly relevant. Machine Learning or AI [1] and in particular self-organising and clustering algorithms, form the basis for the theory. In particular, the relation between these algorithms and the human brain [11][6][7][8][9], are considered and the author's work is related much more to AI than to the topic of Physics.

## 3    Self-Organising System

Machine Learning [1] can be broadly categorised into supervised and unsupervised learning. With supervised learning there is a teacher, and the procedure is more accurate, because the system can skew its weight values to fit data that it is explicitly told about. With unsupervised learning, the system also has to decide on broad organisation details and it can only make a general assessment there, which often relates to recognising centres of pattern concentrations and assigning category classifications to those centres. There are however, many algorithms that can do this and the more sophisticated ones also make use of distributions and other hidden trends in the data. A new self-organising algorithm that will be described in a different paper [7], is based quite closely to some brain principles derived from earlier work. Those principles are described next and then it is shown how they evolved into a measure of cohesion and finally the entropy equation of this paper. The theory starts with Hebb's law [11], which is: 'neurons that fire together wire together'. This is a well-known doctrine that was discovered by Hebb and it helps to explain how the brain orgnaises itself. Neurons in close proximity to each other will form links if they consistently fire at the same time and so there is an attraction between neuron signals that causes this to happen. From this doctrine, it is possible to make some other statements, such as:

- The shorter distance in the brain fires first. This is because it is more economic.
- Equal distances in the brain fire first. This is because they are more synchronized.

These two statements relate to the idea that the brain tries to economise its energy consumption as much as possible [9], or reduce its energy state, or entropy. They also lead to the idea of 'mean distance' and 'variance' as being important measurements when determining how the brain might organise itself. It is also known that the brain likes to be fully connected ([2] and the author's related work). With regard to pattern formations, this is the most economic way to determine that a set of nodes belong together, as they quickly reinforce that group over other nodes. These 3 rules, if you like, have led to the new self-organising algorithm that will be published in the paper [7]. It was then realised however, that the same rules had been used as part of an earlier cohesion equation [8] that was supposed to determine if pattern formations belonged together or separately. The cohesion equation is described below:

Var = (1.0 - (standard deviation / mean local count))         Equation 1
CFp = (mean local count / mean global count)  or  mean R      Equation 2
Cohp = (Var * CFp)                                            Equation 3

Where:
Var = variance standard deviation that uses the average local count.
CFp = count difference to scale the variance by.
Cohp = cohesive value for the whole pattern.

The equation was derived by considering reinforcement counts, but it is not very effective for trying to determine cluster groups as part of a self-organising system. It may be the case that it works better when pattern structures are already determined and the decision is about consistency over those structures. Ignoring the specifics about the different types of count, the essential properties are given in Equation 3, which is the 'mean' relationship with the 'variance' as the key measure.

## 4   Entropy

Section 3 has described the essential properties of the brain model that have gone into some new equations for clustering patterns. Cohesion is a measure of consistency and so it would be possible to replace cohesion with the more general measure of entropy. The reinforcement counts can also be generalised to some mean value and so the entropy equation becomes:

Entropy = (mean * variance)                                                                    Equation 4

Where mean and variance are the standard statistical measurements. What they measure is not defined, but distance is likely to be key. This equation therefore states that: for there to be entropy requires a distance between the entities and also some measure of difference. It also states that as the distance tends to 0 or the variance tends to 0, then so does the entropy, or it achieves maximum order.

Entropy is a measure of disorder in a system, where it is zero when there is complete order. The second law of thermodynamics however states that a system tends to disorder, or maximum entropy and as a result of this, work is done through the creation of the chaos. With relation to the human brain, this measurement might be applied more appropriately to energy than to information. In the paper [6] it is suggested that the wiring between neurons is actually critical for storing information, but that the information held in the wiring may be of a type and not a specific value. The wiring length, for example, would relate to a type of feature and so neurons grouped together in a pattern would represent

DCS	25 July 2020

this feature. Therefore possibly, the brain entropy has more to do with energy, represented by the fixed structure, than with signal instances that flow through the structure.

### 4.1   Entropy in Physics

Looking at the entropy Equation 4, it has exactly the same structure as Einstein's famous equation for relativity: $E = mc^2$. Entropy is also a key consideration for measuring the Universe, especially how it is created or might die. Therefore, this section extends the discussion into that realm and gives a lightweight opinion about how entropy might be able to replace the view of energy, as part of the equations in that domain. To start with, Newton realised that forces were only in existance when objects were moving, and moving relative to each other, so this is a clear indication of variance. He also proposed that gravity was a force. It is still not known exactly what gravity is and this paper does not propose a solution, but theories includes details about waves or particles, forces and electromagnetic waves. Entropic gravity [12] might be interesting. It suggests an emerging, not constant, type of gravity, when the relative position of the two entites is important. If that is the case, then a distance, maybe extending to the centre of the bodies involved, would correlate quite well with relative masses. If comparing to Einstein's equation, distance can somehow replace the idea of mass and variance can replace speed. Speed is different to variance and is the speed of light, which is a constant that does not die in a vacuum. It is simply a very large number that the electromagnetic spectrum can generate from the interation of the electric and the magnetic waves, and the correlation here is a large amount of variance instead. In theory, it is described that the electric forces are created from the interactions between the positively and the negatively charged particles and the magnetic forces are created from the relative movements of these particles. This therefore requires a variation between the entities involved and if the variation did not exist, it is probably the case that the electromagnetic wave would not exist either. If the electric particles cannot move, for example, then there is no magnetic field and the EM wave cannot sustain itself. Therefore, energy requires this disorder to sustain itself.

Then why not consider Black Holes themselves [10]. They are known to be massively dense, but exactly what is going on inside is a bit of a mystery, because it cannot be measured.

Light, for example, cannot escape a black hole after passing the event horizon. It is thought that the force of gravity inside a black hole is so great that something must travel faster than the speed of light to escape it. Black holes also have maximum entropy and trap any matter or energy inside. But there are contradicting theories and so what if there are simply no forces acting inside of the black hole, or the energy is all potential? If there is no work being done inside of the black hole, then light does not have any energy to escape with. If the density of the black hole squeezes the mass closely-enough together, then maybe the particles have no room to move and so they are not able to create any energy from that movement. As a single solid body, it would in fact have maximum order. That would make sense in terms of the entropy equation. It might mean however that the black hole would emit some type of charge, but without work being done, so that it could continue to attract outside bodies.

## 5   Conclusions

This paper describes a short journey, from some very basic brain equations, to entropy and then over to a theory about the universe itself. It is simply a shot in the dark with no claims about any level of expertise. The brain part has already been written about and focuses mostly on energy consumption and synchronization of the firing elements. The equations move from clustering through similarity, to a more global style of entropy, and from information to energy. It may be the case that energy requires disorder to exist (first law of thermodynamics) and this disorder can be measured by something as simple as relative distance and difference. The second law of thermodynamics however, states that a system tends towards chaos and maximum entropy, when it then loses its ability to do more work. An orderly system is therefore at a higher potential state and that potential gets used up when work is done, resulting in the disorder. Something like a snowflake however, shows how a controlled energy input can produce order again. Likewise, if the system is reversed into a zero-entropy state, it would be turned into one with maximum potential, but again unable to do work. Work is done through the energy released by the change in entropy between the two states - where mass cannot be compacted any more and mass cannot dissipate any more. The entropy Equation 4 fits with these ideas and it could be so broad that statements about the forces in black holes can be made. This could have great importance, especially if gravity is charge-related. If it was possible to disrupt the black hole structure, then if the compacted particles started to move again, they might release that huge amount of potential energy, resulting in a big bang. This could also be an interesting source of alternative energy, if very dense material could be created, because it would not produce harmful emissions.

## References

[1] Alpaydin, E. (2020). Introduction to machine learning. MIT press.

[2] Anderson, J.A., Silverstein, J.W., Ritz, S.A. and Jones, R.A. (1977) Distinctive Features, Categorical Perception, and Probability Learning: Some Applications of a Neural Model, Psychological Review, Vol. 84, No. 5.

[3] Einstein A. (1916), Relativity: The Special and General Theory (Translation 1920), New York: H. Holt and Company

[4] Feynman, R.P. (2005). The Feynman Lectures on Physics. 2 (2nd ed.). Addison-Wesley. ISBN 978-0-8053-9065-0.

[5] Goldstein, M. and Inge F. (1993). The Refrigerator and the Universe: Understanding the Laws of Energy. Harvard Univ. Press. ISBN 9780674753259.

[6] Greer, K. (2019). A Concept-Value Network as a Brain Model, available on arXiv at https://arxiv.org/abs/1904.04579.

[7] Greer, K. (2019). A Pattern-Hierarchy Classifier for Reduced Teaching, available on arXiv at https://arxiv.org/abs/1904.07786.

[8] Greer, K. (2017). A Brain-like Cognitive Process with Shared Methods, Int. J. Advanced Intelligence Paradigms (IJAIP), Inderscience, waiting to be published.

[9] Greer, K. (2014). New Ideas for Brain Modelling 2, in: K. Arai et al. (eds.), Intelligent Systems in Science and Information 2014, Studies in Computational Intelligence, Vol. 591, pp. 23 - 39, Springer International Publishing Switzerland, 2015, DOI 10.1007/978-3-319-14654-6_2. Published on arXiv at http://arxiv.org/abs/1408.5490. Extended version of the SAI'14 paper, Arguments for Nested Patterns in Neural Ensembles.

[10] Hawking, S. (1988). A Brief History of Time. Bantam Books. ISBN 978-0-553-38016-3.

[11] Hebb, D.O. (1949). The Organisation of Behaviour.

[12] Verlinde, E.P. (2011). On the origin of gravity and the laws of Newton. J. High Energ. Phys. No. 29. https://doi.org/10.1007/JHEP04(2011)029.